\pdfoutput=1

\documentclass[11pt]{article}

\usepackage[]{acl}
\usepackage{inconsolata}

\usepackage{hyperref}
\usepackage{url}
\usepackage{times}
\usepackage{latexsym}
\usepackage{url}
\usepackage{booktabs}
\usepackage{multirow}
\usepackage{graphicx}
\usepackage{graphicx}
\usepackage{subfig}
\usepackage{txfonts}
\usepackage{pifont}

\usepackage[T1]{fontenc}

\usepackage[utf8]{inputenc}

\usepackage{microtype}

\title{NLP Progress in Indigenous Latin American Languages}

\author{\normalsize Atnafu Lambebo Tonja$^{\Diamondblack,\varheartsuit}$, Fazlourrahman Balouchzahi$^{\Diamondblack}$,  Sabur Butt$^{\clubsuit	}$,  \\
\textbf{\normalsize  Olga Kolesnikova$^{\Diamondblack}$,  Hector Ceballos$^{\clubsuit}$,  Alexander Gelbukh$^{\Diamondblack}$, } \\
\textbf{\normalsize Thamar Solorio$^{\varheartsuit,\spadesuit}$} \\
 $^\Diamondblack$ Instituto Politécnico Nacional, Mexico\\
 $^\clubsuit$ Tecnológico de Monterrey Mexico
\\
  $^\varheartsuit$ MBZUAI, Masdar City, UAE\\
  $^\spadesuit$ University of Houston, Houston, USA
}

\begin{document}
\maketitle
\begin{abstract}
The paper focuses on the marginalization of indigenous language communities in the face of rapid technological advancements. We highlight the cultural richness of these languages and the risk they face of being overlooked in the realm of Natural Language Processing (NLP). We aim to bridge the gap between these communities and researchers, emphasizing the need for inclusive technological advancements that respect indigenous community perspectives. We show the NLP progress of indigenous Latin American languages and the survey that covers the status of indigenous languages in Latin America, their representation in NLP, and the challenges and innovations required for their preservation and development. The paper contributes to the current literature in understanding the need and progress of NLP for indigenous communities of Latin America, specifically low-resource and indigenous communities in general. 
\end{abstract}

\section{Introduction}
In a rapidly changing world with the advent of cutting-edge technologies, the emergence of Artificial Intelligence (AI), and the development of highly intelligent systems, it is essential to draw our attention to communities that seem to dwell in a different realm, detached from the whirlwind of global progress. These individuals, those who do not possess the knowledge or ability to speak the world's dominant languages, risk becoming increasingly marginalized and alienated from the ever-changing global landscape. But who are they, and why are they so crucial to the broader tapestry of human existence? These questions beckon us to delve deeper into the significance of preserving and promoting indigenous languages.

The people who speak indigenous languages represent a rich tapestry of cultural diversity~\cite{nakagawa2021identities}. They are the bearers of unique worldviews, traditions, and ancestral knowledge that have been passed down through generations. These languages are not just a means of communication; they are vessels of history, folklore, and the wisdom of their communities. When we neglect or lose these languages, we forfeit a vital part of our collective human heritage. Each indigenous language holds the key to preserving a distinct cultural identity and the intangible yet invaluable assets that come with it.

Indigenous languages face an even greater degree of underrepresentation within the field of Natural Language Processing (NLP).~\citet{joshi2020state} have highlighted a striking fact: over 88\% of the world's languages, spoken by approximately 1.2 billion people, have been overlooked and continue to be neglected in the realm of language technologies. ~\citet{blasi2022systematic} have further emphasized that while linguistic NLP tasks such as morphology analysis exhibit a more inclusive approach to language diversity, user-facing NLP tasks like machine translation (MT) tend to be less accommodating. In today's information age, NLP techniques have become pervasive in their application on the internet, significantly shaping the content we encounter daily. Consequently, the absence of NLP technology support for endangered languages restricts their visibility to users. This unfortunate situation exacerbates the issue of linguistic marginalization, as regular exposure to a language is pivotal for its continued use and development. Conversely, most NLP research exhibits a bias toward languages with abundant resources, neglecting a wide array of linguistic typologies~\cite{joshi2020state} and often relying on the availability of extensive datasets. Incorporating endangered languages into NLP research can serve to assess the generalizability of NLP models~\cite{bender2011achieving} and promote the pursuit of universal and resource-efficient approaches. The lack of alignment between the under-representation of indigenous languages in NLP and the pressing need for their inclusion motivates the creation of this study. Our aim is to bridge the gap between these Latin American communities and researchers, facilitating a more profound and reciprocal dialogue. By compiling insights, challenges, and innovations related to indigenous languages, we hope to provide researchers with a clearer roadmap, helping them prioritize what is vital and pressing. This paper seeks to offer a comprehensive overview of the current needs in the realm of NLP, ultimately fostering a more inclusive, collaborative approach to technological advancements that respect the unique perspectives and aspirations of indigenous Latin American people. 

We summarize the contributions of this paper as follows:

\begin{itemize}
    \item Report and discussion of the current state-of-the-art NLP research efforts for indigenous Latin American Languages. 
    \item Analysis of challenges and opportunities in contributing to this space.
    \item We provide recommendations for researchers interested in indigenous Latin American Languages based on community feedback.
\end{itemize}

\section{Overview of Indigenous Languages of Latin America}~\label{Indig_overview}
Indigenous peoples, who make up around 5\% of the world's population, collectively preserve more than 7,000 distinct languages, showcasing their remarkable linguistic diversity~\cite{un_indigenous_day}. In Latin America, notable indigenous languages include Quechua, which traces its origins to what is now Ecuador and Peru with around ten million speakers; Guarani, which serves as the official language of Paraguay with over six million speakers; Nahuatl with approximately two million speakers in Mexico, Aymara is spoken by about two million people in Peru, Chile, and Bolivia, and Mapundung (Mapuche), an influential indigenous language with unclear origins in Chile and Argentina~\cite{un_indigenous_day}. 
Indigenous languages hold a significance extending well beyond mere markers of identity or community affiliation. They encapsulate the ethical values derived from ancestral wisdom, fostering a profound connection with the land, and stand as a vital cornerstone for preserving indigenous heritage and nurturing the aspirations of younger generations. This critical situation of indigenous languages is substantiated by data from~\cite{nordhoff2012glottolog}, revealing the existence of around 86 language families and 95 language isolates in the region, with an alarming number of these languages classified as endangered. The imminent threat of extinction primarily stems from state policies. Some governments actively engage in eradicating these languages, including extreme measures such as criminalizing their use, reminiscent of the historical context during the early colonial era in the Americas. Furthermore, the plight of indigenous languages is exacerbated by the denial of the existence of indigenous peoples in certain nations. This denial relegates their languages to mere dialects, affording them less recognition in comparison to national languages, thereby accelerating their decline~\citep{degawan2019indigenous}.
\begin{figure*}[h!]
    \centering
\includegraphics[width=\textwidth]{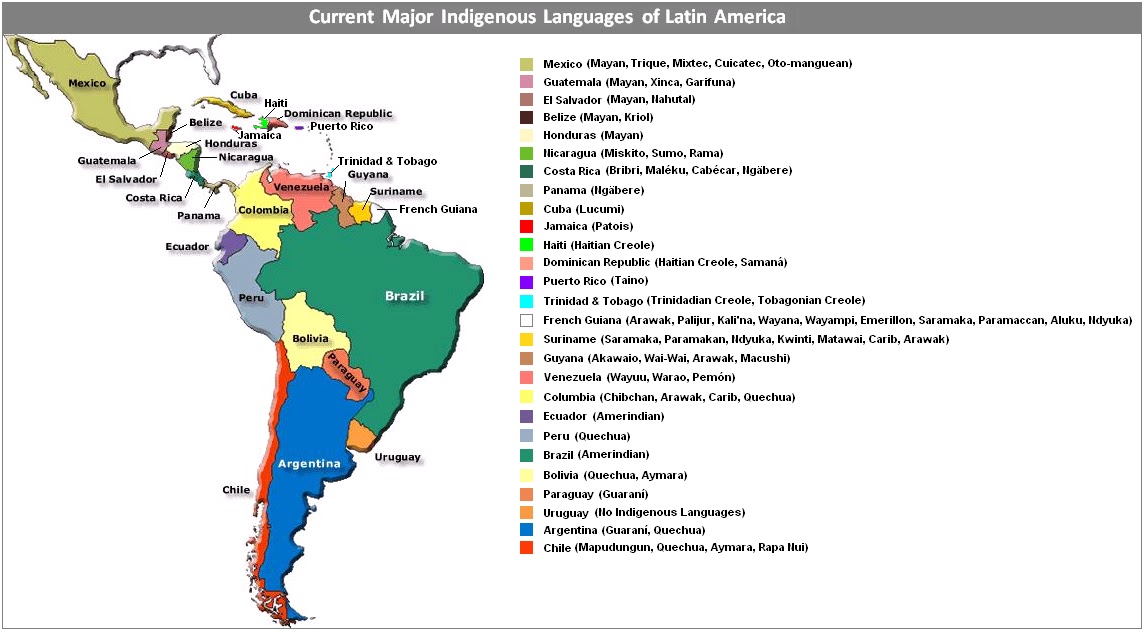}
    \caption{Modern Indigenous Languages in Latin America. Source \url{https://adockrill.blogspot.com/2012/05/map-of-contemporary-latin-america.html}}
    \label{fig:languagemapamerica}
\end{figure*}

The linguistic panorama in Latin America showcases an extraordinary assortment of indigenous languages, encompassing diverse language families, isolates, and unclassified linguistic forms~\citep{campbell2020contact}. With an estimated tally of approximately 650 languages, including both existing and extinct varieties, this region stands as a testament to the multifaceted cultural and historical legacy it holds~\cite{indigenous_languages_americas_families}. Distinctive language families like Mayan, Na-Dené, Algic, Arawakan, Tupian, Quechuan, Cariban, and Uto-Aztecan, among others, contribute significantly to this intricate linguistic fabric, each carrying profound significance in reflecting unique socio-cultural heritages throughout Latin America. This linguistic richness permeates seamlessly across the entire region, presenting a shared legacy of indigenous linguistic diversity. Current efforts aimed at documenting and preserving these languages emphasize their crucial role in upholding the deep cultural heritage and identity of indigenous communities across Latin America~\cite{indigenous_languages_americas_families}.

The map presented in Figure~\ref{fig:languagemapamerica} offers a visual representation of the indigenous languages that continue to be actively spoken across Latin America~\cite{dockrill2023}. A recent report from Statista~\cite{statista2023} highlighted that approximately 7.5\% of the Latin American population use native or indigenous languages as their mother tongues. 
\section{Overview of efforts in NLP research  for indigenous Latin American languages} ~\label{Section: NLP_progress}
This section discusses the progress of NLP works for Indigenous Latin American languages. We used ACL Anthology\footnote{\url{https://aclanthology.org/}} to search NLP works and searched the literature using keywords: Latin America, indigenous language of Latin America, indigenous language of (*) -- * denotes Latin American languages. 
The number of languages depicted in the figure per country shows that most languages are left behind in NLP research in each country. For example, in Mexico, the government recognizes 68 indigenous languages. However, only around half (22) of the indigenous languages are represented in NLP research; in Peru, over 70 indigenous languages are spoken, but only 12 languages are present in NLP research. 
\begin{figure}
    \centering
\includegraphics[width= 0.5 \textwidth]{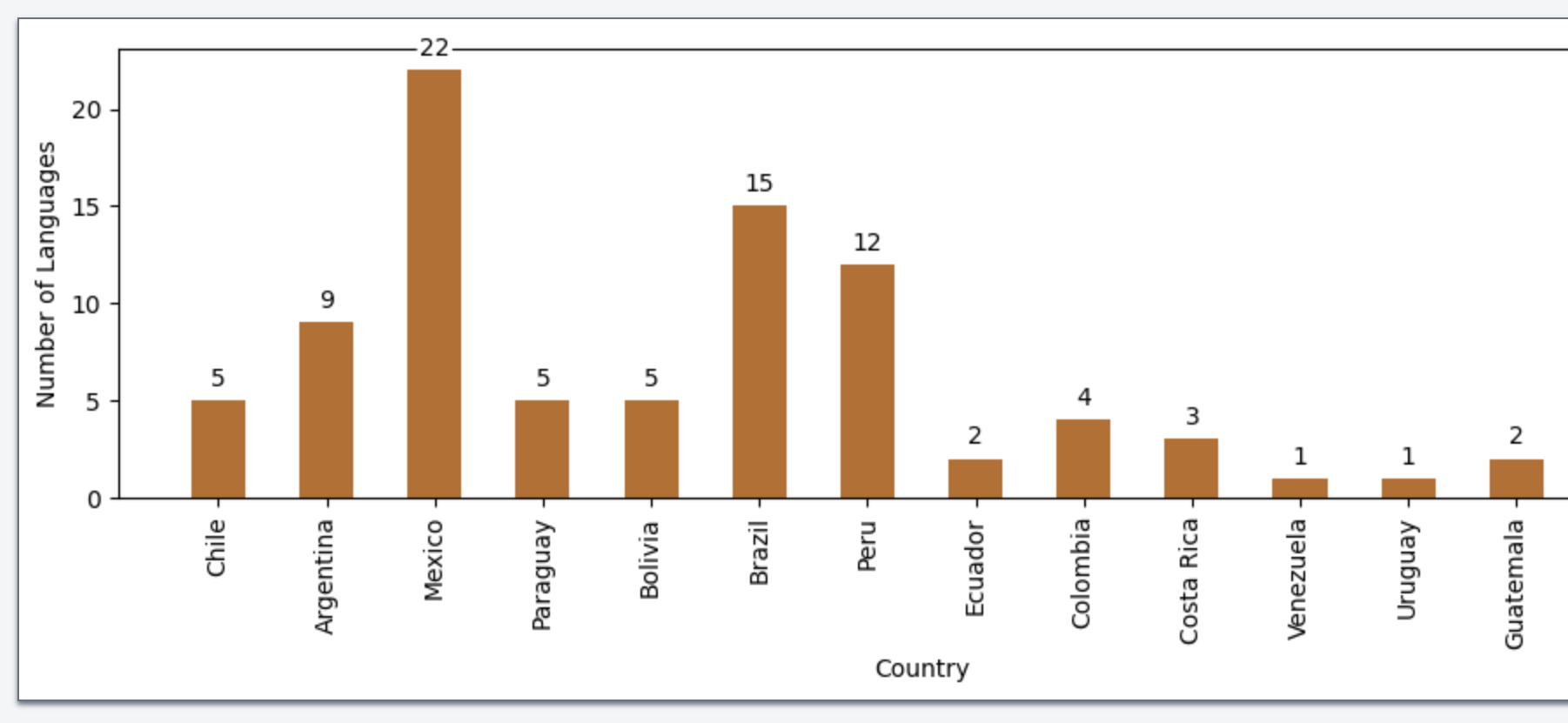}
    \caption{Number of indigenous languages present in NLP research per country}
    \label{fig:no_of_lang_vs_country}
\end{figure}
\subsection{Number of indigenous languages present in NLP research per country}
Figure~\ref{fig:no_of_lang_vs_country} shows the number of indigenous languages included in NLP research/work per country.  Out of 33 countries in the Americas, we found NLP works for indigenous languages spoken in 13 countries. Seven of these countries are represented with five or more languages, and the rest have 1 to 4 languages. As shown in the figure, Mexico has the highest number of representations, with 22 languages in total. Following Mexico, Brazil and Peru have 15 and 12 languages represented in NLP research. Argentina comes next with nine languages, followed by Chile, Paraguay, and Bolivia, each with five languages, and the rest of the countries have 1 to 4 languages.
\begin{figure*}
    \centering
\includegraphics[width=\textwidth]{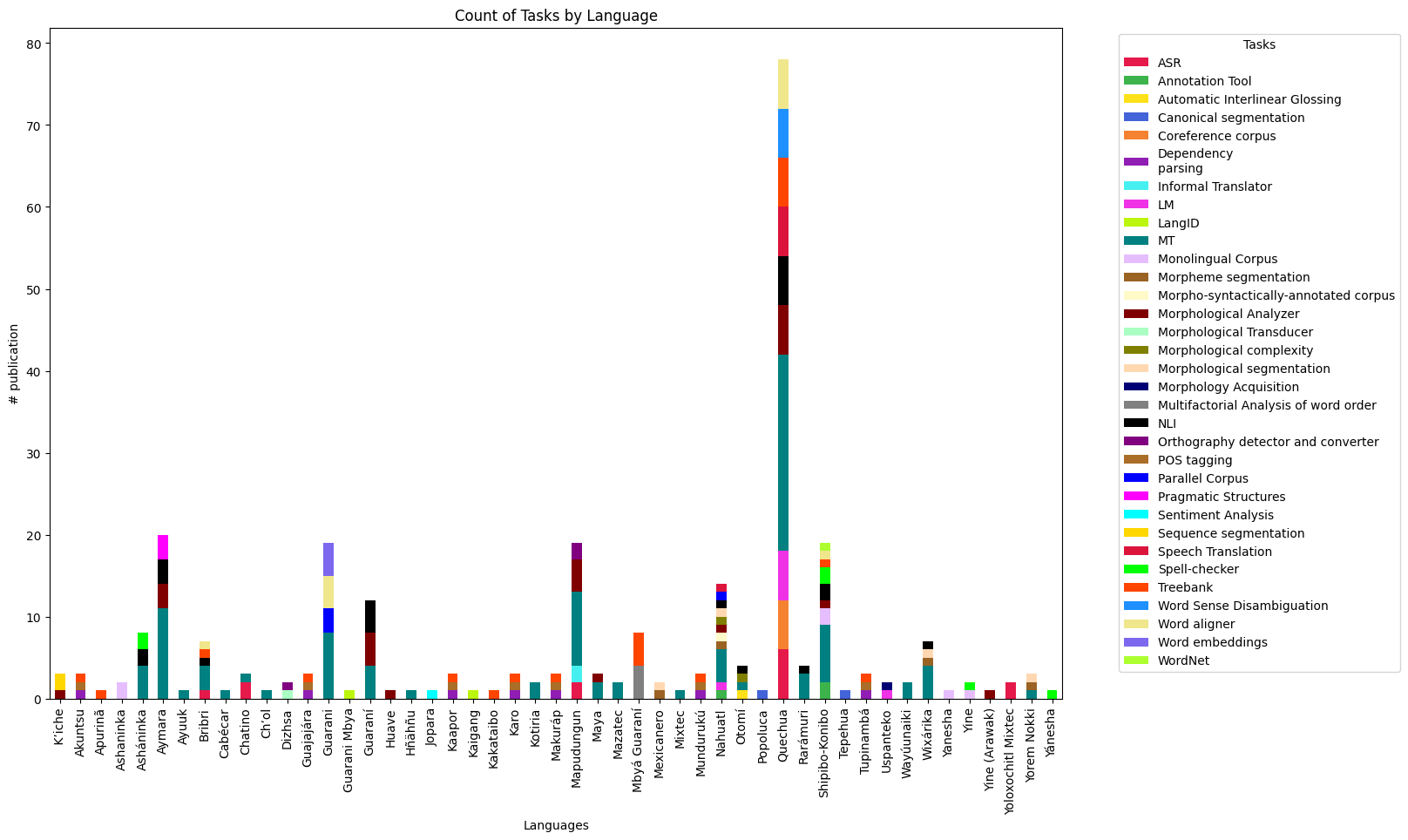}
    \caption{Number of publications per language vs tasks. We did not include publications, tasks, and languages from shared tasks like AmericasNLP in these statistics. }
    \label{fig:Number_of_publication_Vs_Languages}
\end{figure*}
\subsection{Number of publications per language vs task}
Figure~\ref{fig:Number_of_publication_Vs_Languages} illustrates the number of publications available for each language vs tasks. From the figure, we observe that Quechua has more research and publications in different downstream NLP tasks, while languages like Shipibo-Konibo, Nahuatl, Aymara, and Guarani also have a good representation, covering more than 2 NLP tasks. Still, the majority of the languages present in papers only have one publication or NLP task. This clearly shows the under-representation of indigenous Latin American languages in NLP research. This highlights a critical gap in linguistic diversity in NLP research in indigenous languages of Latin America, emphasizing the need for increased focus and resources towards underrepresented languages to foster a more inclusive and equitable technological advancement.  
\subsection{Total number of publications per task}
\begin{figure}
    \centering
    \includegraphics[width= 0.5\textwidth]{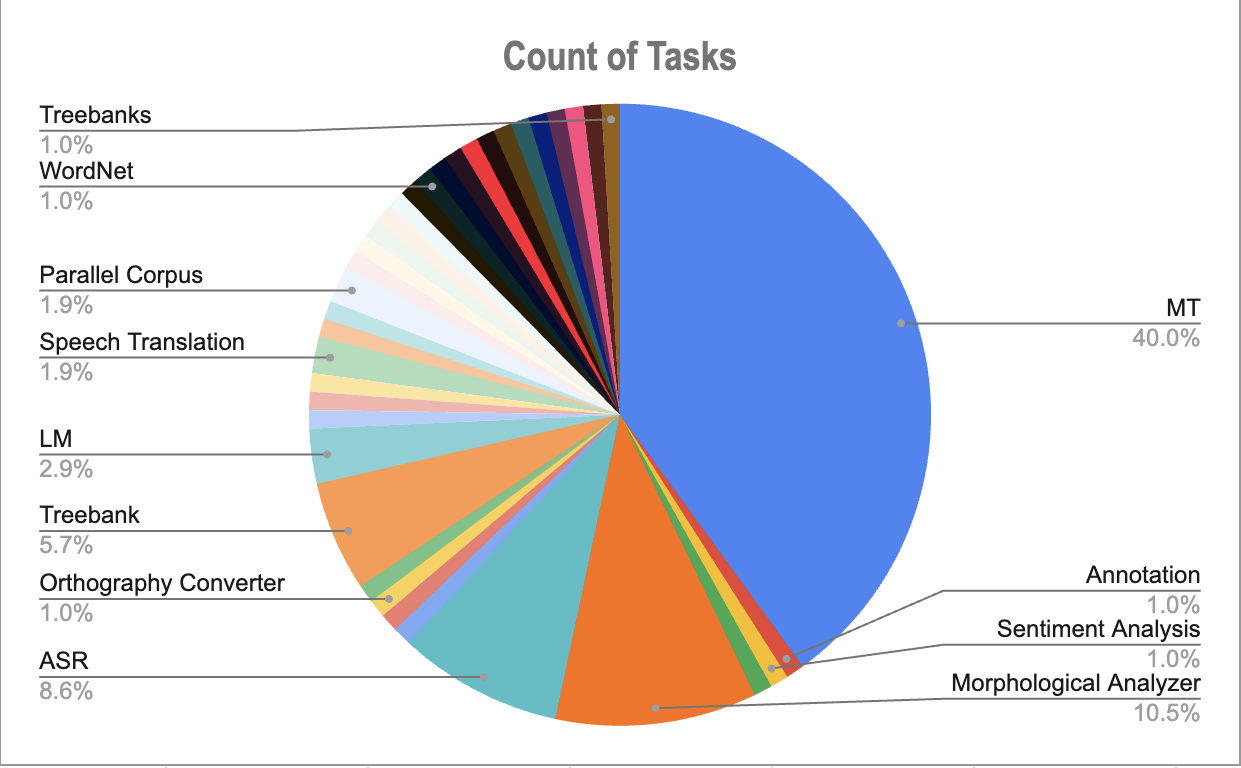}
    \caption{Total number of publications per task}
    \label{fig:No_of_publication_vs_task}
\end{figure}
Figure \ref{fig:No_of_publication_vs_task} depicts the overall number of publications per task. As we can see from the figure, the machine translation task (MT) is the most researched NLP task, with $\approx 40\%$ of overall publications, while Treebank, Morphological Analyzer, and Automatic Speech Recognition (ASR) have more than $\approx 5\%$ publications.  

We also found 12  MT shared task papers submitted in shared tasks organized by AmercasNLP in 2021, 2022, and 2023 that propose different methods to tackle MT problems in diverse indigenous Latin American languages. 
\subsection{Publication per year, venues and type }
\begin{figure*}[ht]
    \centering
    \subfloat[\centering Venues ]{{\includegraphics[width=5cm]{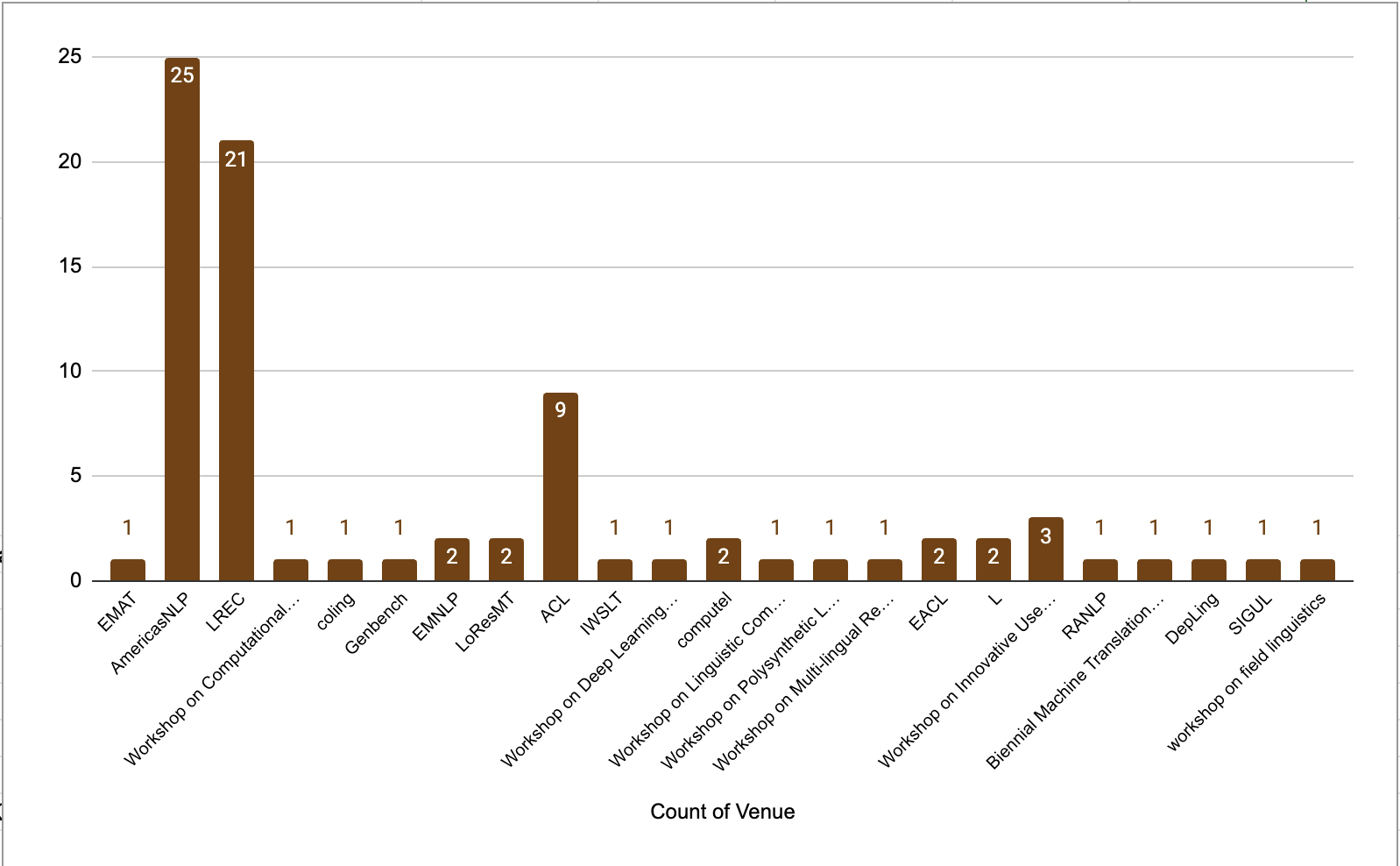} }} 
    \qquad
     \subfloat[\centering Publication type]{{\includegraphics[width=3cm]{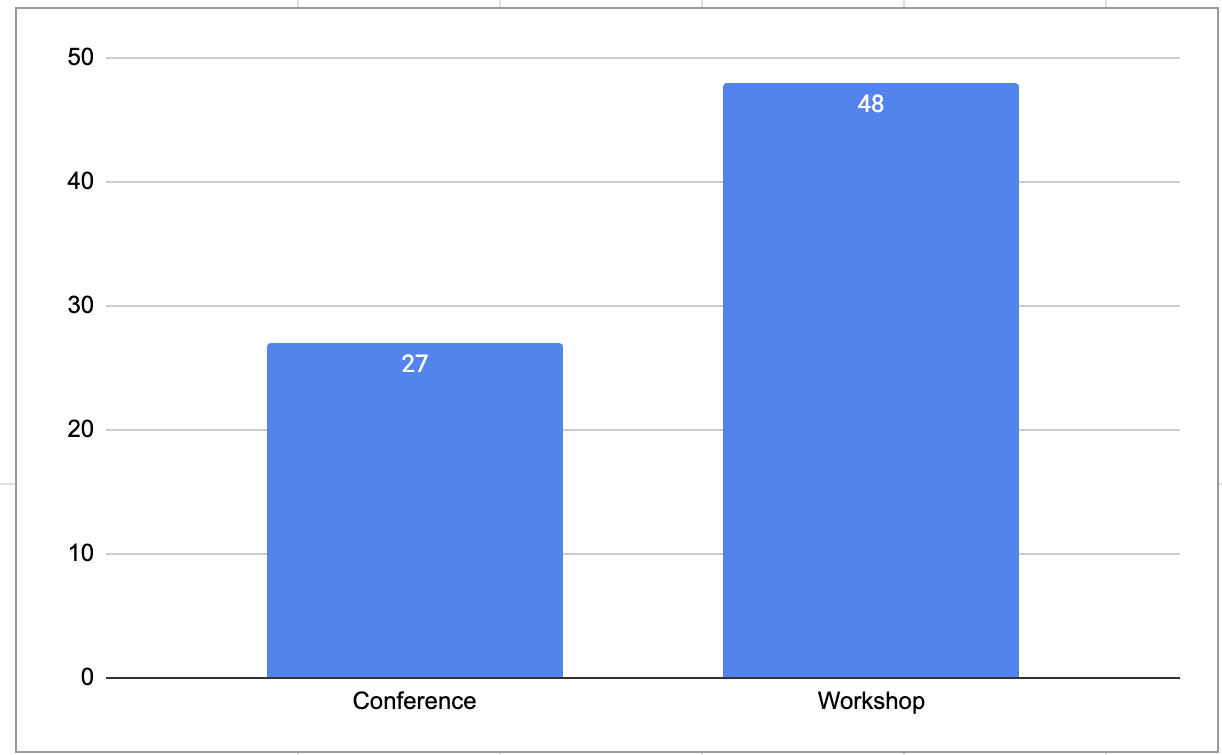} }}
     \qquad
    \subfloat[\centering Year vs Publication]{{\includegraphics[width=3cm]{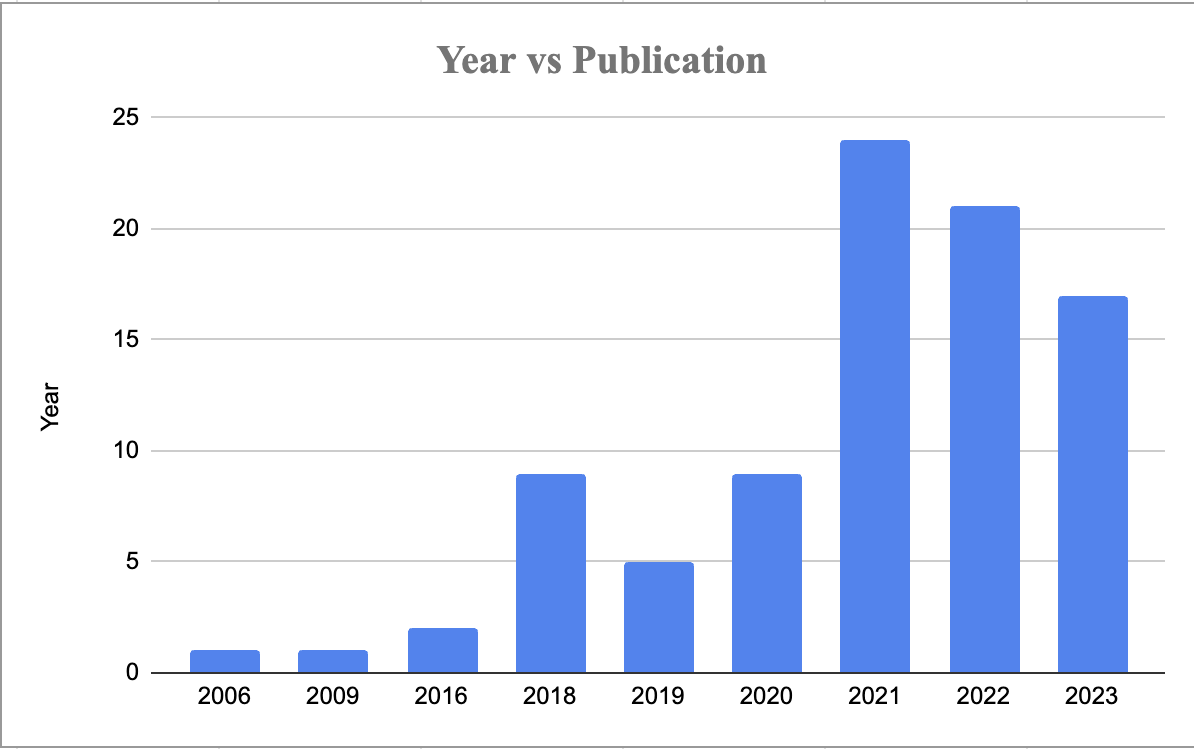} }}
    
    \caption{Publication per year, venues, and publication types}
    \label{fig:pub_year_venue}
\end{figure*}

Figure \ref{fig:pub_year_venue} shows the overall research efforts of NLP publications for Latin American languages. As we can see from the figure, workshops are the dominant type of publication venue, surpassing conferences, and when we observe the specific venues, \textbf{AmericasNLP} and \textbf{LREC} are the leading venues over others. When we look at the years and number of publications, we see that the number of publications is increasing over time, and we can see dramatic changes starting in 2021. This clearly shows that the \textbf{AmericasNLP} workshop, which started in 2021, increased opportunities for researchers working in Latin American languages to publish their work. We observed more publications in LREC, a conference dedicated to publishing resources for different languages. We can observe the publications in top NLP conferences like ACL, EMNLP, EACL, etc. 


\section{Reflection from the community and researchers}~\label{Section: NLP_progress_why_less} 
To better understand the challenges of working in indigenous languages and recommend a suitable research direction for interested researchers, we conducted a survey with students, indigenous language speakers, and researchers working on indigenous languages.

\textbf{Survey Design --} Our survey focuses on two groups of people: (1) researchers who are familiar with NLP or/and working on indigenous Latin American languages and (2) the indigenous language community.   

\textbf{Participant Selection --} We contacted researchers (including professors) and students from two research institutions in Latin American countries
and researchers from regional NLP groups.
We explained the aim of the survey and its output to them, and we received 27 responses in total from them.  To survey indigenous language speakers and the general public, we used ClickWorker\footnote{\url{clickworker.com}}, a survey outsourcing platform to hire participants from all Latin American countries. We received 350 responses from them; all participants hired by ClickWorker are paid for their work.

\begin{figure*}[ht]
    \centering
    \subfloat[\centering Age distribution ]{{\includegraphics[width=3cm]{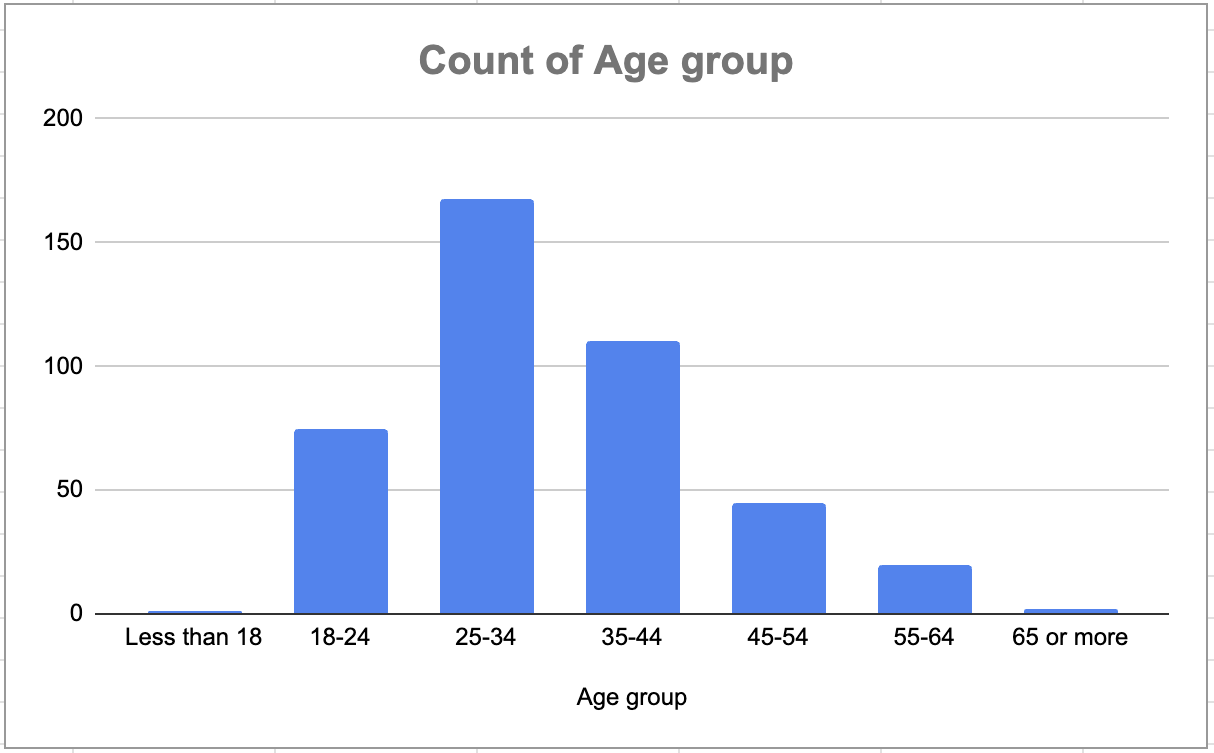} }} 
    \qquad
     \subfloat[\centering Country distribution]{{\includegraphics[width=3cm]{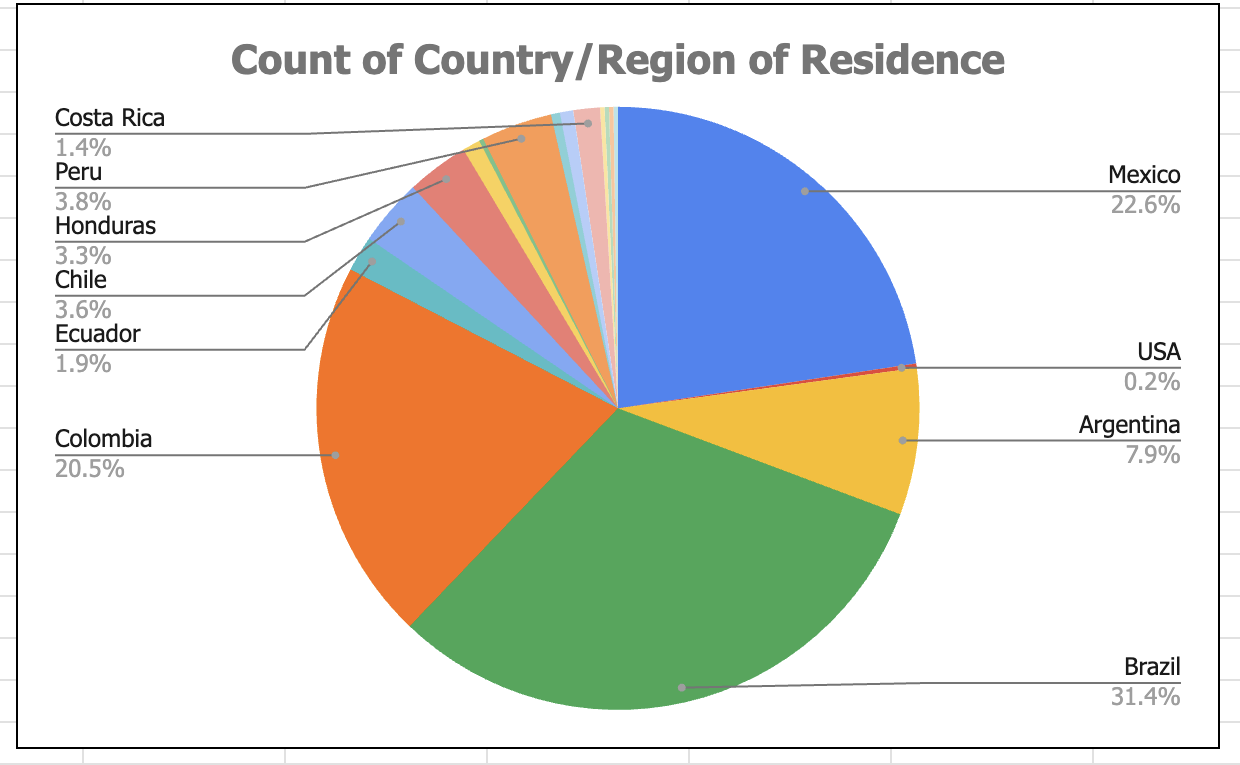} }}
     \qquad
    \subfloat[\centering Education statistics]{{\includegraphics[width=3cm]{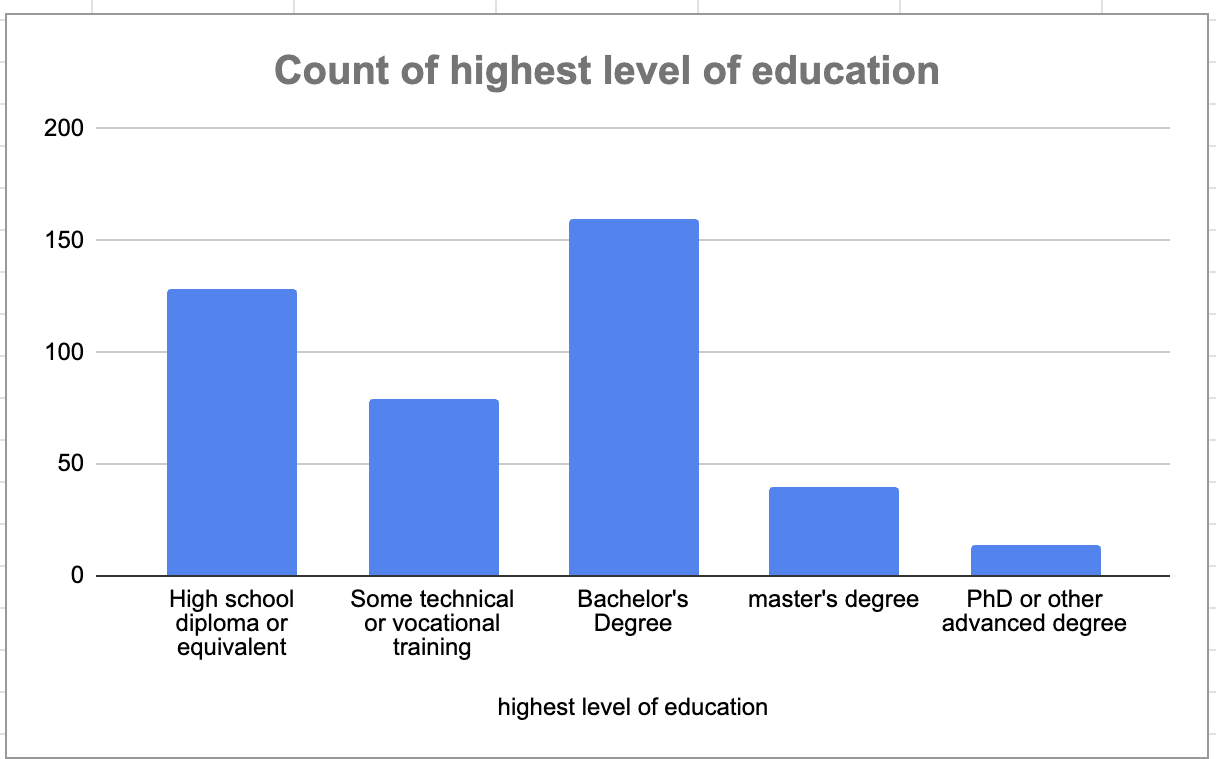} }}
    \qquad
    \subfloat[\centering Gender distribution]{{\includegraphics[width=3cm]{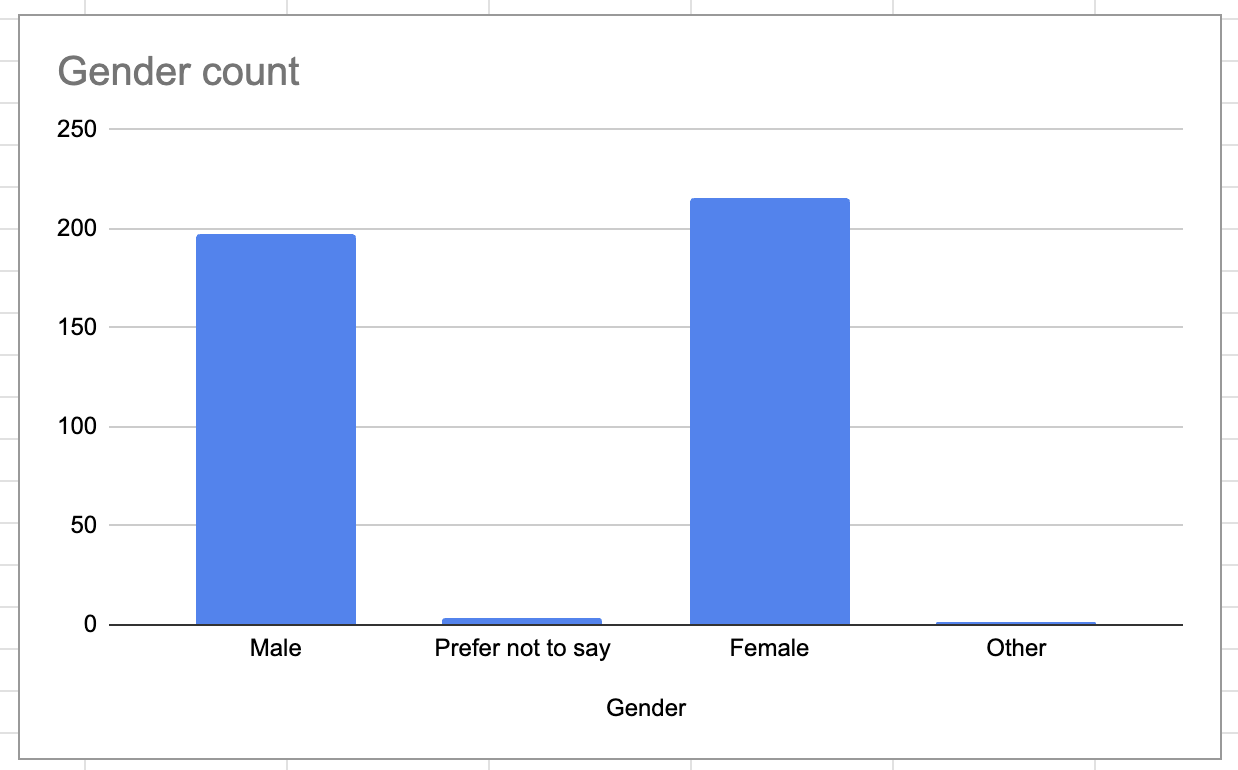} }}
    \qquad
    \caption{Demographic statistics of the participants}
    \label{fig:survey_stats}
\end{figure*}

\textbf{Survey Questions --} Our survey consists of 19 questions, from which nine questions ask about participants' general information like Gender, Age, Education, Country, indigenous languages they speak/are familiar with, and their occupation/experience. The remaining questions are about indigenous languages and NLP, challenges and needs, collaboration, and support. All the responses are anonymous and will not cause any harm to the participants. We received board approval from our institution before conducting the study. Figure \ref{fig:survey_stats} shows statistics of the survey responses.

We categorized our survey questions into four main topics as follows:-
\textbf{1)} What challenges do researchers and indigenous communities encounter in NLP research for indigenous languages? \textbf{2)} What does the community need to save/preserve their languages? \textbf{3)} What opportunities do we currently have to develop different NLP tools for these languages? \textbf{4)} What future direction should be considered to increase the research works for these languages?

    
\textbf{What challenges do researchers and indigenous communities encounter in NLP research for indigenous languages?}
    
NLP for indigenous languages involves developing computational tools to process and understand these languages, which often have rich oral histories but limited written records, at the intersection of technology, linguistics, and cultural preservation \cite{ward2018qualitative}. Preserving and revitalizing native languages is crucial for indigenous communities as it helps them maintain their cultural identity \cite{marmion2014community}. Moreover, it allows researchers to broaden the scope of NLP to include more diverse and inclusive languages, moving beyond the currently dominant ones in the field \cite{joshi2020state}. 

Regarding the challenges faced by researchers and the indigenous community, we received the following responses: \textit{Researchers -} 
highlighted the following key issues:-  \textbf{\textit{lack of access to resources}, \textit{lack of effort from the scientific community}, \textit{lack of indigenous language speaker involvement in the creation of the gold standard dataset}}. In contrast, respondents from \textit{indigenous community -} emphasized the following challenges: \textbf{\textit{lack of inclusion of the indigenous community in research process}, \textit{little interest from the government}, \textit{the greatest inclusion of anglicisms and new technology.}}

Based on the responses received, it is evident that there are multiple challenges at play when it comes to NLP research in indigenous Latin American languages. We observed different responses regarding challenges from researchers and communities. Researchers are more focused on the availability of resources and scientific community efforts, whereas responses from the indigenous communities are more focused on including the indigenous community in the research process and the government's interest. The foremost problem is the lack of resources, which implies that these communities need additional tools and support to become active participants in scientific research. This problem is further aggravated by the scientific community's need to be more proactive in incorporating indigenous knowledge and perspectives, suggesting an imbalanced approach to the research and development process.

Moreover, the under-representation of indigenous language speakers in creating key datasets is a significant barrier to inclusivity. This exclusion restricts the diversity and richness of the data and undermines the relevance and usefulness of research outcomes for indigenous populations.

These issues are further exacerbated by the lack of interest shown by governmental bodies, which indicates more extensive systemic neglect. The prevalence of Western concepts, the use of Anglicicisms, and new technologies suggest a cultural overshadowing. It is necessary to give more consideration and importance to indigenous ways of knowing and doing. In science and technology, indigenous voices and needs are marginalized, leading to a homogenized and less inclusive approach to research and technological advancement.

        


 \textbf{What does the community need to save/preserve their languages?} 

Linguists advocate for the use of one's native language as an essential initial measure in safeguarding any linguistic heritage. Without such efforts, indigenous languages risk gradual erosion under the influence of dominant group languages. A critical inquiry arises: are our technological strides fostering increased utilization of native languages, thereby empowering speakers of these languages? Or are they reinforcing the dominance of widely spoken languages? Essentially, do these advancements contribute to the preservation of languages?



The insights from survey respondents highlighted the distinct yet interconnected roles of stakeholders, such as technology companies, government entities, and academic institutions. Technology companies emerge as crucial contributors, providing essential financial and technical resources. These resources can be subsidized for the indigenous communities to facilitate better outreach and access to technology. Governments, as outlined by respondents, are urged to play a proactive role by implementing incentive policies. The \textit{\textbf{role of academic institutions}} is also pivotal in facilitating collaborative research. By bringing together a diverse range of experts and actively involving indigenous communities, these institutions address the unique challenges NLP poses for indigenous languages. Developing educational resources and training programs should also emerge as a strategy to promote a better understanding of NLP technologies within indigenous communities. Investors are encouraged to \textit{\textbf{formulate and invest in AI applications that actively contribute to the preservation of indigenous culture}}, stressing the inclusion of the indigenous community in the process and having them guide the priorities on what is needed for their communities and avoiding acculturation. In alliance with educational institutions, governments are advised to focus on providing broad education for indigenous communities, allocating resources to include them in technological advancements, thus preventing the disappearance of native languages. Institutions are positioned as key players in this comprehensive approach, with a responsibility to facilitate new technologies, provide economic assistance, and offer training and support for the holistic development and preservation of indigenous languages and cultures in Latin America. 

Central to these suggestions is the imperative to \textit{\textbf{involve members from indigenous communities}} at all stages of development, ensuring representation and cultural sensitivity. The call for representative data collection highlighted the importance of \textit{\textbf{diverse datasets}} to train NLP models effectively. \textit{\textbf{Transparency, documentation, and ethical audits}} were recommended practices by the respondents to address algorithmic bias and promote accountability. The survey emphasized respecting copyright and cultural rights and the development of translation tools that consider linguistic and cultural nuances.

\textit{\textbf{Cultural sensitivity and understanding}} are paramount in integrating NLP technologies within indigenous contexts. Recommendations included avoiding generalizations, conducting extensive studies on indigenous peoples to inform technology development, and better understanding their culture for a respectful approach. Respondents emphasized profound respect for indigenous peoples' customs, thoughts, and traditions, advocating for active listening and keeping NLP applications separate from indigenous culture to prevent misappropriation. Additionally, \textit{\textbf{effective communication and collaboration}} with indigenous communities were highlighted as crucial. This involved direct contact with leaders and representatives, leveraging online resources for independent dissemination, and actively involving communities in projects and initiatives. Open-source initiatives were encouraged for transparency, and integrating indigenous community members in projects was recommended for genuine inclusion. 

\textit{\textbf{Respect and privacy}} were repeated subjects in the responses, emphasizing the importance of establishing agreements that honored cultural boundaries and avoided privacy invasion. Recommendations included respecting people's decisions, adapting to community preferences, and conducting activities at the community's pace and within their cultural context. This collective emphasis on respect and cultural understanding forms a crucial foundation for ethical and meaningful engagements with indigenous communities. The respondents also addressed the importance of \textit{\textbf{government and policy involvement}} for the welfare of indigenous communities. Recommendations included government intervention, human rights organizations, public policies, and union work. The comments stressed the necessity of government support for effective and sustainable initiatives benefiting indigenous communities.

A multifaceted approach to \textit{\textbf{education and awareness}} is recommended, with an emphasis on comprehensive awareness programs, education, changing paradigms, and targeted initiatives to disseminate accurate information. Teaching the language and creating awareness through meaningful work have been suggested as effective ways to instill mutual respect among individuals, with a specific focus on educational workshops as practical tools for achieving these goals. Practical steps and actions were proposed, including hiring native speakers, direct contact with communities, on-site research, and putting indigenous languages on platforms. The implementation of compensation systems, limits, and sanctions were suggested for fair collaboration and accountability. Adapting NLP to the reality of indigenous communities was recommended for relevant applications, supporting and preserving indigenous languages and cultures.

The survey comments also highlighted strategic recommendations for \textit{\textbf{research and development}}, emphasizing increased investment, detailed studies, regularizing efforts with protocols, and implementing oversight and monitoring systems. Improving resources, expanding language databases, and investing in the preservation of languages and dialects were key strategies. The survey underscored the importance of contributing resources for comprehensive studies, utilizing advanced technologies, and staying current through updated research and surveys. The overarching theme was a strategic and well-supported effort to advance research and understanding of indigenous languages and cultures.

\textbf{What future direction should be considered to increase the research works for these languages?}~\label{Future_direction}

Despite the challenges discussed earlier in this section, promising opportunities exist to promote NLP research and tools for these languages. Based on Figure~\ref{fig:No_of_publication_vs_task}, it is evident that a substantial portion of researchers in these languages are exploring MT. However, there remains a crucial need for researchers to focus on tasks that have received less attention. This requires a thorough examination of the current status and necessary advancements for each language, including the identification of tasks present or absent within their respective linguistic contexts. Drawing from previous efforts to support extremely low-resource languages~\cite{maldonadotowards, llitjos2005building, gasser2011computational, mager2021findings}, we present a discussion on certain tasks and assess the languages in which these tasks can be effectively implemented.

\begin{itemize}
\item Machine Translation (MT) - Most of NLP research in indigenous Latin American languages centers on MT, as indicated in Figure~\ref{fig:Number_of_publication_Vs_Languages}. While certain languages like Quechua, Shipibo-Konibo, Nahuatl, Aymara, Guarani, Mapudungun, and Wixarika have garnered considerable attention in MT research, numerous others, such as Yánesha, Yoloxochitl Mixtec, Tepehua, Otomi, Popoluca, remain relatively unexplored, lacking both research initiatives and associated tools within this domain.
    \item Morphology - is crucial for understanding the linguistic intricacies of Indigenous Latin American languages, which are esteemed for their cultural richness and offer valuable insights into human linguistic diversity. Despite their significance, these languages often lack resources for morphological research and technological development. Researchers have initiated various morphology-related tasks, such as morpheme segmentation, annotated corpora, analyzers, and transducers, particularly focusing on languages like Quechua and Nahuatl. However, there remains a significant gap in tools and research for many other languages, as illustrated in Figure~\ref{fig:Number_of_publication_Vs_Languages}. Some of these languages are Apurinã, Asháninka, Bribr, Cabécar, Jopara, etc.

    \item Speech Recognition and Translation - Figure~\ref{fig:Number_of_publication_Vs_Languages} shows that several efforts have been made for ASR and speech translation for Quechua, Yoloxochitl Mixtec, and Wixarika, but the majority of languages such as Raramuri, k'iche, Akuntsu, Apurinã, Aymara, etc. still need tools and researches in this domain.

    \item Named Entity Recognition (NER) and Part-of-Speech (POS) tagging - represent two essential NLP tasks for which there is a notable scarcity of tools and research dedicated to indigenous languages. As illustrated in Figure~\ref{fig:Number_of_publication_Vs_Languages}, the majority of languages, such as Quechua, Shipibo-Konibo, Guarani, Kakataibo, Jopara, among others, suffer from a lack of resources for these specific tasks.

    \item Treebank - is a parsed text corpus that annotates syntactic or semantic sentence structure \cite{taylor2003penn}, are fundamental for the development of various NLP tools in any low-resource languages. Lexical databases are crucial in enhancing NLP systems by offering a foundation for understanding a language's lexicon. This necessity is highlighted by the statistics presented in Figure~\ref{fig:Number_of_publication_Vs_Languages}, where Shipibo-Konibo stands as the sole exception, underscoring the need for such resources for these tasks.

    \item Language Identification (LI) - is another primary requirement for facilitating the development of language technologies for these languages. Figure~\ref{fig:Number_of_publication_Vs_Languages} shows that LI can be an ideal initiative in NLP tasks for languages such as Mazatec, Maya, Mixtec, Kaapor, Popoluca, and many more.

    \item Natural Language Inference (NLI), as depicted in Figure~\ref{fig:Number_of_publication_Vs_Languages}, along with machine translation (MT) and speech recognition tasks, possesses some resources, albeit limited, in languages like Quechua, Asháninka, Aymara, Bribri, Guarani, Nahuatl, and Shipibo-Konibo, while other languages such as Mapudungun, Mexicanero, Yorem Nokki, Yine, Popoluca, among others, lack any resources for NLI.

    \item Word embeddings - as indicated by the statistics in Figure~\ref{fig:Number_of_publication_Vs_Languages}, are notably absent for these languages, with the exception of Guarani, despite their significance and the demonstrated value as essential tools for any language.
\end{itemize}
\textbf{Country vs frequency of responses}

\begin{figure*}
    \centering
    \includegraphics[width=\textwidth]{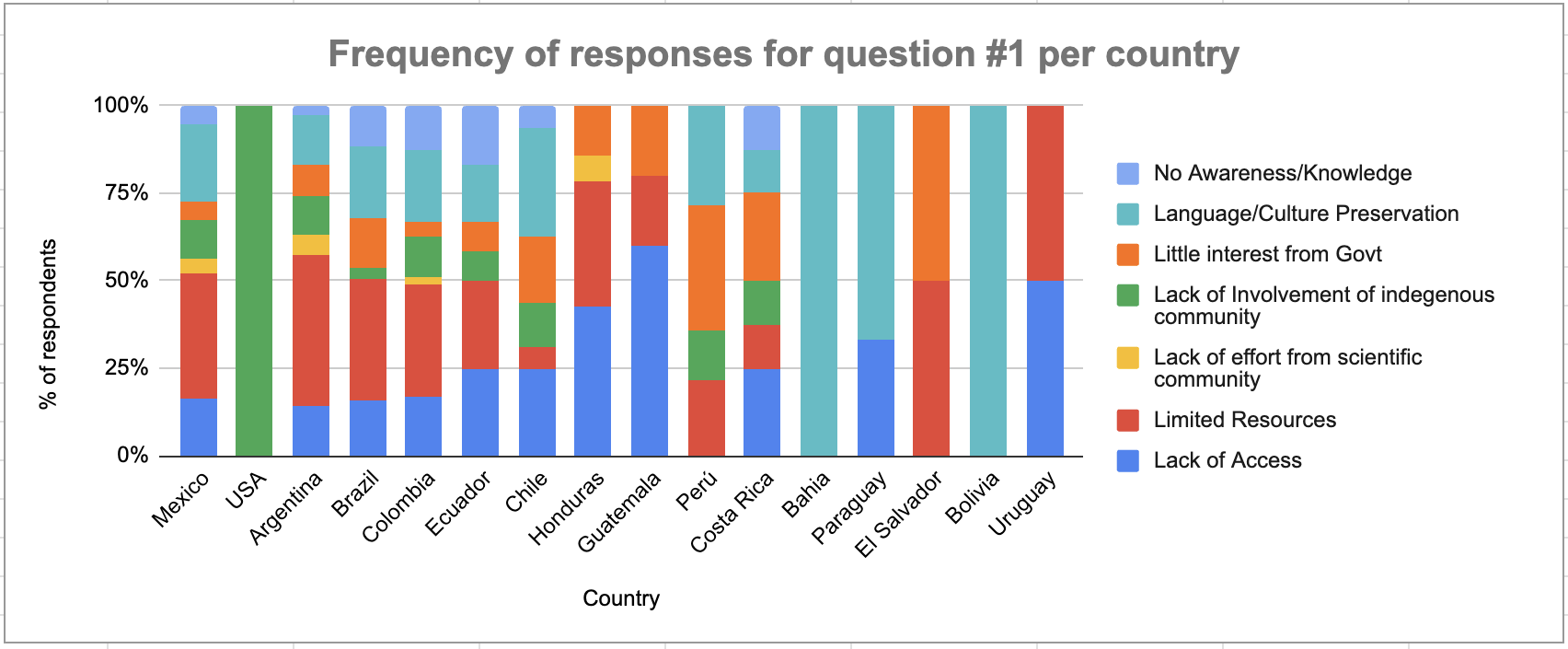}
    \caption{Frequency of response distribution across different Latin American countries}
    \label{fig:freq_cunt}
\end{figure*}

Figure \ref{fig:freq_cunt} shows the frequency of responses vs the country of respondents for the first question: \textit{What challenges do researchers and indigenous communities encounter in NLP research for indigenous languages?}  
As we can observe from the figure, the responses from participants from different countries reflect common challenges. When we see the frequency of the responses that \textbf{Limited resource, Lack of access, Language/culture preservation and Lack of interest from government} have a higher frequency than the others. 

\textbf{Limited Resources} - Latin American indigenous communities face a multitude of challenges in adopting NLP and AI, primarily due to limited resources. Financial and technological constraints hinder their access to advanced technologies, while the lack of digitized linguistic resources threatens language preservation. Limited data availability and difficulties in accessing modern technology further impede progress. Economic factors, including insufficient investment, pose significant barriers to technology adoption.	

\textbf{Lack of Access} - manifests in various forms, including limited resources for technological infrastructure and education. Many communities struggle with inadequate internet connectivity, hindering their ability to utilize modern technologies effectively. Additionally, geographic isolation and economic constraints exacerbate the issue, preventing access to essential tools and resources. Lack of education and training further compound the problem, as does the preservation of cultural identity, which can sometimes conflict with the adoption of new technologies.  We added a detailed explanation about this response in Section \ref{App-2}.    
						
\section{Takeaways}
\subsection{Opportunities for NLP researchers and indigenous communities }
As discussed in Section \ref{Section: NLP_progress}, we observed promising works progress for Latin American indigenous languages, especially a dramatic increase in publications for those languages in recent years. 
We observed the following takeaways from Section \ref{Section: NLP_progress}.  

\textbf{Indigenous languages --} We observe the promising number of publications for languages like Nahuatl, Quechua, Shipibo-Konibo, Bribri, Mapudungun, Aymar, and Wixarika in different NLP domains.

\textbf{NLP tasks --} We observe different works done for indigenous languages of Latin America; from those NLP tasks, MT is one of the leading tasks for those languages. We also observe high-level NLP tasks like the Pre-trained language model for Quechua and low-level NLP tasks like spell checker, morphological analyzer, and Treebank for most languages. 

\textbf{Community impact --} We also observe the impact of communities like AmericasNLP by creating environments for researchers to present and publish their works, which Figure \ref{fig:pub_year_venue} notices. 




\subsection{Community reflection}
Feedback from researchers and indigenous communities pointed out interesting points regarding challenges, community needs, and future directions. Researchers working on these languages should include the community in every research process to gain the community's trust and obey the community's customs.

\section{Conclusion}
In this work, we explore the research progress in indigenous Latin American languages, and we conduct a survey study to identify the challenges when working on these languages, as well as the needs and future directions of the research in NLP research. We hope this study will show some direction for researchers interested in indigenous Latin American languages and low-resource indigenous languages in general.

\section{Limitations}
This study is limited to Latin American languages and NLP research efforts on those languages. The analysis focused only on showing the research efforts in the area of NLP; we did not include a detailed literature review for downstream NLP tasks. For the overview, we only used ACL Anthology papers. The summary of the responses from the community and researchers may or may not be generalized to a broader society or languages.  As discussed in the paper, the motivation of this study is to show the works that have been done for indigenous Latin American languages and to understand the challenges and needs of the community and the scientific community.
\section{Ethical consideration}
We conducted the survey after board approval from our institution and agreement from the respondents, who agreed to participate in the study. Participants also consented to the use of their data and responses for research work and publication. All the survey participants are anonymous, and we did not include questions that expose their anonymity. The Clickworker platform compensates all community members who participated in this survey. 

\section{Acknowledgments}
The authors would like to acknowledge the financial and technical support of Writing Lab, Institute for the Future of Education, Tecnologico de Monterrey, Mexico, in the production of this work. 

\bibliography{anthology,custom}
\bibliographystyle{acl_natbib}

\appendix
\section{Papers we found from ACL Anthology}
\cite{zheng2021low,levin2009adaptable,tonja2023parallel,tonja2023enhancing,vazquez2021helsinki, mager2023neural, chiruzzo2022jojajovai,tyers2023codex,aguero2021logistical,de2023four,cavalin2023understanding,monson2006building,zevallos2022huqariq,mager2018challenges,parida2021open,pugh2023developing,chen2022improving,pugh2022universal,duan2019resource,nagoudi2021indt5,schwartz2022primum,feldman2020neural,graichen2023enriching,washington2021towards, kiss2019word,zariquiey2022cld2,cavar2016endangered,thomas2019universal,ginn2023robust,cordova2021toward,moon2009unsupervised,adams2019evaluating,rios2012tree,montoya2019continuous,knowles2021nrc,kuznetsova2021finite,pendas2023neural,gongora2021experiments,ebrahimi2021americasnli,ortega2023quespa, himoro2022preliminary, zevallos2022introducing,ortega2018using,martinez2020cplm,tyers2023towards,marquez2021ayuuk,shi2021highland,gutierrez2018comparing,stap2023chatgpt,martinez2021automatic,pugh2023finite,pugh2021investigating,bollmann2021moses,gutierrez2016axolotl,mager2018lost,kann2018fortification,mager2023ethical,kann2018fortification,amith2021end,solano2021explicit,jones2023talamt,tan2023few,shi2021leveraging,bustamante2020no,gongora2022can,mager2020tackling,mager2022bpe,mercado2018chanot,oncevay2021peru,oncevay2022schaman,maguino2018wordnet,torres2021representation,zariquiey2022building,ronald2018morphological,alva2017spell,moreno2021repu,vasquez2018toward,gow2023sheffield,ahmed2023enhancing,ahumada2022educational,downey2021multilingual,pankratz2021qxoref,rudnick2011towards,ebrahimi2023meeting,chen2021morphologically,blum2022evaluating,homola2013pragmatic,rodriguez2022tupian,kann2022machine,rueter2021apurin,coto2021towards,de2010computational,chiruzzo2020development,tyers2021corpus,tyers2021survey}
\section{Country vs frequency of responses explanation} \label{App-2}

\textbf{No awareness/Knowledge}: This lack of awareness manifests in various forms, including limited understanding of the technologies themselves, uncertainty about the availability and accuracy of datasets for indigenous languages, and a general lack of information about AI and NLP applications tailored to their needs. Additionally, there is a sense that the lack of knowledge leads to a lack of interest and devaluation of indigenous culture and people. Some attribute this lack of knowledge to educational barriers, linguistic challenges, and cultural resistance to change. 	

\textbf{Language/Culture Preservation}: Language preservation emerges as a primary concern, as many indigenous languages lack representation in digital tools and face the risk of being lost or altered over time. Cultural preservation is also highlighted, with a focus on maintaining traditions, customs, and ancestral knowledge. The lack of options and resources in indigenous languages exacerbates the challenge, as does the resistance to modern technology due to its potential impact on linguistic identity. Despite these obstacles, these communities have a strong desire to preserve their languages and cultures, indicating the need for tailored solutions that respect their unique identities and address their specific linguistic and cultural needs.					

\textbf{Little interest from Govt}: This limited interest is evidenced by the absence of public policies and initiatives aimed at utilizing these technologies for the benefit of indigenous communities. Additionally, there is a lack of resources and support from governments for the development and preservation of indigenous languages and cultures. Institutional marginalization, discrimination, and lack of recognition exacerbate the challenges faced by these communities, leading to social isolation and hindering their access to essential resources and opportunities.								

\textbf{Lack of involvement of indigenous community} : This lack of involvement stems from various factors, including resistance to change and new technologies within indigenous communities, limited education and knowledge about technological aids, and cultural barriers. Additionally, there is a perception of exclusion and discrimination, with indigenous languages and cultures often overlooked or undervalued in the digital world. Furthermore, there is a need for greater inclusion and integration of indigenous perspectives and languages in the development of NLP and AI technologies to ensure that these tools adequately serve the needs of indigenous communities and contribute to preserving their languages and cultures.	

\textbf{Lack of effort from scientific community}:  This lack of effort is characterized by insufficient research and interest in developing solutions tailored to the needs of indigenous languages and cultures. Many feel that the scientific community should prioritize adapting NLP and AI technologies to native languages to preserve cultural identity and ensure inclusivity.								

\textbf{Limited Resources}: Latin American indigenous communities face a multitude of challenges in adopting NLP and AI, primarily due to limited resources. Financial and technological constraints hinder their access to advanced technologies, while the lack of digitized linguistic resources threatens language preservation. Limited data availability and difficulties in accessing modern technology further impede progress. Economic factors, including insufficient investment, pose significant barriers to technology adoption.	

\textbf{Lack of Access}: This lack of access manifests in various forms, including limited resources for technological infrastructure and education. Many communities struggle with inadequate internet connectivity, hindering their ability to utilize modern technologies effectively. Additionally, geographic isolation and economic constraints exacerbate the issue, preventing access to essential tools and resources. Lack of education and training further compound the problem, as does the preservation of cultural identity, which can sometimes conflict with the adoption of new technologies. 		
\end{document}